\documentclass[review]{elsarticle}

\usepackage{lineno,hyperref,amsmath}
\usepackage[ruled]{algorithm2e}
\usepackage{graphicx}
\usepackage{caption}
\usepackage{float}
\usepackage{subcaption}
\usepackage{bm}
\usepackage{xcolor}
\usepackage{booktabs}
\usepackage{amssymb}

\modulolinenumbers[5]

\journal{Journal of Mathematical Imaging and Vision}

\DeclareMathOperator*{\argmin}{arg\,min}

\begin{document}

\begin{frontmatter}

\title{\textit{Using the Split Bregman Algorithm to Solve the Self-repelling Snake Model}}

\author[a]{Huizhu Pan}\corref{mycorrespondingauthor}
\cortext[mycorrespondingauthor]{Corresponding author}
\ead{huizhu.pan@postgrad.curtin.edu.au}
\author[b]{Jintao Song}
\author[a]{Wanquan Liu}
\author[a]{Ling Li}
\author[a]{Guanglu Zhou}
\author[a]{Lu Tan}
\author[a]{Shichu Chen}

\address[a]{School of Electrical Engineering Computing and Mathematical Sciences, 
Curtin University, Perth WA 6102, Australia}

\address[b]{College of Computer Science and Technology, Qingdao University, Qingdao Shandong 266071, China}

\begin{abstract}

%Although this solution is stable, the memory requirement grows quickly as the image size increases. 
Preserving contour topology during image segmentation is useful in many practical scenarios. By keeping the contours isomorphic, it is possible to prevent over-segmentation and under-segmentation, as well as to adhere to given topologies. The Self-repelling Snake model (SR) is a variational model that preserves contour topology by combining a non-local repulsion term with the geodesic active contour model (GAC). The SR is traditionally solved using the additive operator splitting (AOS) scheme. 
In our paper, we propose an alternative solution to the SR using the Split Bregman method. Our algorithm breaks the problem down into simpler sub-problems to use lower-order evolution equations and a simple projection scheme rather than re-initialization. The sub-problems can be solved via fast Fourier transform (FFT) or an approximate soft thresholding formula which maintains stability, shortening the convergence time, and reduces the memory requirement. The Split Bregman and AOS algorithms are compared theoretically and experimentally.

\end{abstract}

\begin{keyword}
\texttt Self-repelling snake\sep Topology-preserving segmentation\sep Split Bregman algorithm\sep Alternating direction optimization\sep Variational method.

\end{keyword}

\end{frontmatter}

\section{Introduction}

Topology preservation in image segmentation is an external constraint to discourage changes in the topology of the segmentation contour. It is typically applied in problems where the object topology is known a priori. One example is in medical image analysis where the segmentation of the brain cortical surface must produce results consistent with the real-world brain cortical structure \cite{41}. Another example is the segmentation of objects with complicated interiors, noises, or occlusions, where a topological constraint can be used to prevent over-segmentation, i.e., the forming of "holes" due to image complexity \cite{19}, or under-segmentation, i.e., when the contours of separate objects merge. Much active research is undergone in the area, such as image segmentation and registration using the Beltrami representation of shapes \cite{1} and non-local shape descriptors \cite{2,3}, multi-label image segmentation with preserved topology \cite{4}, and min-cut/max-flow segmentation using topology priors \cite{5}. 

Since the problem of topology preservation can be intuitively linked to the process of contour evolution, many active contour models \cite{6,7,8} have been proposed for it. In these models, the contour is affected by various forces until it converges to the final segmentation result. To preserve topology during the contour evolution process, a constraint term is usually added to the variational formulation which prevents the contour from self-intersecting, i.e., merging or splitting. For example, Han et al. \cite{10} proposed a simple-point detection scheme in an implicit level set framework in 2003. Meanwhile, Cecil et al. \cite{12} monitored the changes in the Jacobian of the level set. In 2005, Alexandrov et al. \cite{13} recast the topology preservation problem to a shape optimization problem of the level sets, where narrow bands around the segmentation contours are discouraged from overlapping. Sundaramoorthi and Yezzi \cite{14} proposed an approach based on knot energy minimization to realize the same effect. Rochery et al. \cite{15} used a similar idea while introducing a non-local regularization term, which was applied in the tracking of long thin objects in remote sensing images. Building on the previous ideas, the self-repelling snake model (SR) was proposed by Le Guyader et al. in 2008 \cite{16}. The SR uses an implicit level set representation for the curve and adds a non-local repulsion term to the classic geodesic active contour model (GAC)\cite{8}. In the follow-up work \cite{17}, the short time existence/uniqueness and Lipschitz regularity property of the SR model were studied. Later, \cite{3} successfully extended the SR model to non-local topology-preserving segmentation-guided registration. Attempts have also been made \cite{19} to combine the SR with the region-based Chan-Vese model, though a direct combination proved less successful than the original SR.

The SR model has intuitive and straightforward geometric interpretations, but its non-local term leads to complications in the numerical implementation. Explicit iterative solutions are unstable and require small time steps, leading to low computational efficiency. To the best of our knowledge, the SR model has always been solved through the additive operator splitting (AOS) \cite{ AOS_old, 48, AOS_Lu} strategy in a semi-implicit way. The AOS strategy is able to solve multidimensional equations as one-dimensional equations which promotes parallelization.
In \cite{48}, arithmetic averaging was replaced by harmonic averaging in the calculation of the discrete geodesic curvature term. With stable the semi-implicit iterations and the fast Thomas algorithm to solve tridiagonal linear systems, the AOS scheme is both reliable and efficient. However, the memory requirements of the coefficient matrices are still considerable and the discretization of geodesic curvature is strenuous to implement. In this paper, we propose an alternative solution using the Split Bregman method to formulate a more concise algorithm which requires less memory, costs less time per iteration, and converges faster.

The Split Bregman algorithm was first proposed in computer vision by Goldstein and Osher \cite{32} for the total variation model (TV) for image restoration. By introducing splitting variables and iterative parameters, it transforms the original constrained minimization problems into simpler sub-problems that can be solved alternatively. The Split Bregman algorithm is shown to be equivalent to the Alternating Direction Method of Multipliers (ADMM) \cite{33} and the Augmented Lagrangian Method (ALM) \cite{34} in a convex setting. In this paper, we introduce an intermediate variable to split the original problem into two sub-problems, which turns a second-order optimization problem into two first-order ones. The two sub-problems can be solved by the Fast Fourier Transform (FFT) method and an approximate generalized soft thresholding formula, respectively, ensuring efficiency and reliability without complicated discretization schemes and the hefty memory requirement. We also replaced the re-initialization of the signed distance function with a simple projection scheme. As a result, the optimization of the level set function is even more simplified. In addition, to address some problems arising from the Split Bregman solution, we replaced the Heaviside representation of the level set in \cite{16} with one that performed better in our algorithm.

The paper is organized as follows. In section 2, we review and provide some intuition to the original SR model. In section 3, we design the Split Bregman algorithm for the SR model and derive the Euler-Lagrange equations or gradient descent equations for the sub-problems. In section 4, the discretization schemes for the sub-problems are presented for the alternating iterative optimization. In section 5, we provide some numerical examples and comparisons of results. Finally, we draw conclusions in section 6.

\section{The Original Self-Repelling Snake Model and the AOS}

The original SR model as proposed in \cite{16} is an edge-based segmentation model based on the GAC \cite{8}. It adopts the variational level set formulation \cite{42}, where the segmentation contour is implicitly represented as the zero level line of a signed distance function \cite{11}. An energy functional is minimized until convergence is reached and the segmentation contour is obtained. The energy functional comprises three terms, two of which are taken from the GAC model and contribute to edge detection and the balloon force respectively, while the last one accounts for the self-repulsion of contour as it approaches itself. 

The definition of the SR model is as follows. Let $f(x):\Omega\rightarrow R$ be a scalar value image, $x\in\Omega$, and $\Omega$ is the domain of the image. The standard edge detect function $g(x)\in[\,0,1]\,$ is given by

\begin{equation}
g(x)=\frac{1}{1+\rho|\nabla(G_\sigma\ast f)|^s},
\end{equation}

\noindent where $s=1$ or 2, $\rho$ is a scaling parameter, and $G_\sigma$ denotes a Gaussian convolution of the image with a standard deviation of $\sigma$. The object boundary $C$ is represented by the zero set of a level set function $\phi$,

\begin{equation}
C=\{x\in\Omega|\phi(x)=0\}.
\end{equation}

The level set function $\phi$ is defined as a signed distance function, such that,

\begin{equation}
\phi(x)=\begin{cases}
    -d(x,C) & x\text{ } inside\text{ } C \\
    0   &  x\in C \\
    d(x,C) & x\text{ } outside\text{ } C\end{cases},
\end{equation}

\noindent where $d(x,C)$ is the Euclidean distance between point $x$ and contour $C$. As a signed distance function, $\phi$ satisfies the constraint condition below, i.e. the Eikonal equation,

\begin{equation}
|\nabla\phi|=1.
\end{equation}

To represent the image area and contour, we use the Heaviside function $H(\phi)$ and Dirac function $\delta(\phi)$. Since the original Heaviside function is discontinuous and therefore not differentiable, we adopt the regularization schemes below \cite{42},

\begin{equation}
H_{\varepsilon}(\phi) = 
\begin{cases}
%\begin{array}{cc}
{ \frac{1}{2} \left( 1 + \frac{\phi}{\varepsilon} 
+ \frac{1}{\pi} \sin \left( \frac{\pi \phi}{\varepsilon} \right)\right)} 
& 
{|\phi| \leq \varepsilon}
\\ 
\hfil 1 & \hfil \phi > \varepsilon
\\ 
\hfil 0 & \hfil \phi<-\varepsilon
%\end{array}
\end{cases},
\end{equation}

\begin{equation}
\delta_{\varepsilon}(\phi)=
%\left\{\begin{array}{cc}
\begin{cases}
{\frac{1}{2 \varepsilon}\left(1+\cos \left(\frac{\pi \phi}{\varepsilon}\right)\right)} & {|\phi| \leq \varepsilon} 
\\
\hfil 0 & {|\phi|>\varepsilon}
%\end{array}.
\end{cases}.
\end{equation}

These regularization schemes are different from the ones in the original model in \cite{16}. Here, $\varepsilon$ does not regularize the entire image domain, which improves stability of edge-based models. The effect is more apparent in our Split Bregman algorithm, as we we will discuss in Section 3.

Given the above, the energy functional $E(\phi)$ of the SR model can be written as

\begin{equation}
E(\phi)=\gamma E_g(\phi)+\alpha E_a(\phi)+\beta E_r(\phi),
\end{equation}

\noindent where $\gamma$, $\alpha$, $\beta$ are penalty parameters that balance three terms.

\begin{equation}
E_g(\phi)=\int_\Omega g(x)|\nabla H_\varepsilon(\phi(x))| dx = \int_\Omega g(x)|\nabla \phi(x)|\delta_\varepsilon(\phi(x)) dx.
\end{equation}

$E_g(\phi)$ is the geodesic length of the contour. The total variation of the Heaviside function, or the total length of the contour, is weighted by the edge detector in (1).

\begin{equation} 
E_a(\phi)=\int_\Omega g(x)(1-H_\varepsilon(\phi(x)))dx.
\end{equation}

$E_a(\phi)$ is the closed area of the contour also weighted by the edge detector. It acts as a balloon force that pushes the segmentation contour over weak edges \cite{7} .

\begin{equation}
E_r(\phi)=-\int_\Omega\int_\Omega e^{-\frac{|x-y|^2}{d^2}}(\nabla\phi(x)\cdot\nabla\phi(y))h_\varepsilon(\phi(x))h_\varepsilon(\phi(y))dxdy.
\end{equation}

$E_r(\phi)$ describes the self-repulsion of the contour \cite{16}. $e^{-\frac{|x-y|^2}{d^2}}$ measures the nearness of the two points $x$ and $y$, e.g. the further away the points the smaller the repulsion. In (10), $h_\varepsilon(\phi(x))$ and $h_\varepsilon(\phi(y))$ denote the narrow bands around the points $x$ and $y$, where,

\begin{equation} 
h_\varepsilon(\phi(x))=H_\varepsilon(\phi(x)+l)(1-H_\varepsilon(\phi(x)-l)),
\end{equation}
\begin{equation} 
h_\varepsilon(\phi(y))=H_\varepsilon(\phi(y)+l)(1-H_\varepsilon(\phi(y)-l)).
\end{equation}

When the points $x$ and $y$ are further than distance $l$ from the contour, $h_{\varepsilon}(\phi(x))h_{\varepsilon}(\phi(y))\rightarrow0$. This causes the points outside the narrow bands to be largely unaffected by repulsion. For $-\nabla\phi(x)\cdot\nabla\phi(y)$, if the outwards unit normal vectors to the level lines passing through $x$ and $y$ have opposite directions, i.e., the contours passing through $x$ and $y$ are merging or splitting, then the functional approaches the maximum value. Thus, the minimization of $E_r(\phi)$ prevents the self-intersection of the contour.

Given the energy functional (7) and the constraint (4) , the variational formulation for SR is
\begin{equation}
    \begin{cases}
    \min\limits_{\phi} E(\phi)=\gamma E_g(\phi)+\alpha E_a(\phi)+\beta E_r(\phi) \\
    \text{s.t. }|\nabla\phi|=1\end{cases},
\end{equation}

\noindent and the evolution equation of $\phi(x)$ derived from $E_g(\phi)$ and $E_a(\phi)$ is

\begin{equation}
\frac{\partial\phi(x,t)}{\partial t}=\delta_\varepsilon(\phi(x,t))(\gamma\nabla\cdot(g(x)\frac{\nabla\phi(x,t)}{|\nabla\phi(x,t)|})+\alpha g(x)),
\end{equation}

\noindent where

\begin{equation}\label{geo_c}
\nabla\cdot(g(x)\frac{\nabla\phi(x,t)}{|\nabla\phi(x,t)|})=\nabla g(x)\cdot\frac{\nabla\phi(x,t)}{|\nabla\phi(x,t)|}+g(x)\nabla\cdot\frac{\nabla\phi(x,t)}{|\nabla\phi(x,t)|}.
\end{equation}

(\ref{geo_c}) is the geodesic curvature that shifts the contour towards the edges detected by $g(x)$. In the image areas with near-uniform intensity, $\nabla g(x)\rightarrow 0$, $g(x)=1$. Since $\nabla\cdot(g(x)\frac{\nabla\phi(x,t)}{|\nabla\phi(x,t)|}) \rightarrow 0$ in those areas, the geodesic curvature term has little effect and the balloon force $\alpha g(x)$ dominates.

Lastly, the evolution equation that can be derived from the repulsion term is

\begin{equation}
\frac{\partial\phi(x,t)}{\partial t}= \frac{4\beta}{d^2} h_\varepsilon(\phi(x,t))\int_\Omega e^{-\frac{|x-y|^2}{d^2}} ((x-y)\cdot\nabla\phi(y,t))h_\varepsilon(\phi(y,t)) dy,
\end{equation}

To summarize, by applying variational methods to the three energy terms and substituting $\delta_\varepsilon(\phi(x))$ with $|\nabla\phi(x)|$, the following evolution equations can be derived

\begin{equation}
    \begin{cases}
    \frac{\partial\phi(x,t)}{\partial t}=|\nabla\phi|(\gamma\nabla\cdot(g(x)\frac{\nabla\phi(x,t)}{|\nabla\phi(x,t)|})+\alpha g(x))\\
    +\frac{4\beta}{d^2} h_\varepsilon(\phi(x,t))\int_\Omega e^{-\frac{|x-y|^2}{d^2}} ((x-y)\cdot\nabla\phi(y,t))h_\varepsilon(\phi(y,t)) dy & x\in\Omega\\
    \phi(x,0)=\phi_0(x) & t=0\\
    \frac{\partial\phi}{\partial\vec{n}}=0 & x\in\partial\Omega\\
    |\nabla\phi|=1
    \end{cases}.
\end{equation}

With regards to the constraint $|\nabla\phi|=1$, the dynamic re-initialization scheme below is adopted in \cite{16},

\begin{equation} 
    \begin{cases}
    \frac{\partial\psi(x,t)}{\partial t}+\sin{(\phi(x))}(|\nabla\psi(x,t)|-1)=0\\
    \psi(x,0)=\phi(x)
    \end{cases}.
\end{equation}

The above is a typical Hamilton-Jacobi equation that can be discretized and solved through an up-wind difference scheme \cite{11}. To solve (17), the original solution adopts the AOS strategy \cite{48, AOS_old}. The first term on the r.h.s. of (17) is discretized with the half-point difference scheme and the harmonic averaging approximation. The next two terms adopt the up-wind scheme. Two semi-implicit schemes are constructed by concatenating the rows and columns of the image respectively \cite{16},

\begin{equation}
\begin{split}
\left( 1-2 \tau A_{l_{1}}\left(\phi^{k}\right)\right) v^{k+1}=\phi^{k}+\tau\left(T^{2}\left(\phi^{k}\right)+T^{3}\left(\phi^{k}\right)\right),
\\
\left(1-2 \tau A_{l_{2}}\left(\phi^{k}\right)\right) w^{k+1}=\phi^{k}+\tau\left(T^{2}\left(\phi^{k}\right)+T^{3}\left(\phi^{k}\right)\right),
\end{split} 
\end{equation}

\noindent where $A_{l_1}, A_{l_2}$ are the two concatenation matrices, $v$ and $w$ are intermediate variables, and $T^2, T^3$ are the up-wind discretizations of the second and third term of the r.h.s. of (17), the formulations of which are omitted here for simplicity. For each $A_{l}\left(l \in\left(l_{1}, l_{2}\right)\right)$,

\begin{equation}
    A_{lij}\left(\phi^{k}\right)=
    \begin{cases}
    \frac{2 \gamma\left|\nabla^{o} \phi_{i}^{k}\right|}{\left(\frac{\left|\nabla^{o} \phi_{i}^{k}\right|}{g_{i}}+\frac{\left|\nabla^{o} \phi_{j}^{k}\right|}{g_{j}}\right)} & j \in N_{l}(i)\\
    -\sum\limits_{m \in N_{l}(i)}\frac{2 \gamma\left|\nabla^{o} \phi_{i}^{k}\right|}{\left(\frac{\left|\nabla^{o} \phi_{i}^{k}\right|}{g_{i}}+\frac{\left|\nabla^{o} \phi_{m}^{k}\right|}{g_{m}}\right)} & j=i\\
    0 & else
    \end{cases},
\end{equation}

\noindent where $i, j$ are two points in the image, $N_l(i)$ is the set of nearest neighbors of $i$ in the matrix $A_l$, $\left|\nabla^{o} \phi_{i}^{k}\right| = \sqrt{\left(\frac{\phi_{i+1, j}-\phi_{i-1, j}}{2}\right)^{2}+\left(\frac{\phi_{i, j+1}-\phi_{i, j+1}}{2}\right)^{2}}$, and $A_l$ is a diagonally dominant tridiagonal matrix. Finally, $\phi^{k+1}$ can be calculated as

\begin{equation}
    \phi^{k+1}=\frac{1}{2}\left(v^{k+1}+w^{k+1}\right).
\end{equation}

\noindent In the last step, (19) is solved via the Thomas algorithm which involves LR decomposition, forward substitution, and backward substitution, with the convergence rate of $O(N)$.

The AOS scheme has several advantages. The semi-implicit formulation is stable and allows for bigger time steps. Furthermore, the algorithm can be executed in parallel along the $l$ directions, which makes it suitable for high dimensional problems. However, the memory requirements of the coefficient matrices are still considerable. Since $i$ and $j$ span the entire image, if $\Omega \in R^{M\times N}$, then $A_l\in R^{(M\times N)\times(M\times N)}$ which means that the memory requirement is quadratic. Additionally, the discretization of the geodesic curvature term is strenuous to implement. 

In the following section, we will propose an alternative solution to the SR with the Split Bregman method that uses more compact intermediate variables, replaces the re-initialization step, and adopts a stable semi-implicit FFT scheme. The aim is to reduce computation time, conserve memory, and maintain stability. Parallelization options will also be discussed in Section 4.

\section{The Split-Bregman Algorithm for the Self-repelling Snake Model}

The Split Bregman method is a fast alternating directional method often used in solving $L^1$-regularized constrained optimization problems \cite{32}. To design the Split Bregman algorithm for (7), we first introduce a splitting variable $\vec{w}=\nabla\phi$ and the Bregman iterator $\vec{b}$. We can re-formulate the energy minimization problem as

\begin{equation}
    \begin{cases}
    (\phi^{k+1},\vec{w}^{k+1})=\argmin\limits_{\phi,\vec{w}}E(\phi,\vec{w})\\
    =\left. \begin{cases}\gamma\int_\Omega g(x)|\vec{w}(x)|\delta_\varepsilon(\phi(x))dx+\alpha\int_\Omega g(x)(1-H_\varepsilon(\phi(x)))dx\\
    -\beta\int_\Omega\int_\Omega e^{-\frac{|x-y|^2}{d^2}}(\vec{w}(x)\cdot\vec{w}(y))h_\varepsilon(\phi(x))h_{\varepsilon}(\phi(y))dxdy\\
    +\frac{\mu}{2}\int_\Omega |\vec{w}(x)-\nabla\phi(x)-\vec{b}^k(x)|^2dx\\
    \end{cases}\right \},\\
    s.t. \text{ }|\vec{w}(x)|=1 
    \end{cases}
\end{equation}

\begin{equation}
    \vec{b}^{k+1}(x)=\vec{b}^k(x)+\nabla\phi^{k+1}(x)-\vec{w}^{k+1}(x),
\end{equation}

\noindent where $\vec{b}^0=\vec{0}$, $\vec{w}^0=\vec{0}$, and $\mu$ is a penalty parameter. The original problem can then be solved as two sub-problems in alternating fashion for loops $k=1,2,...,K$. The sub-problems are,

\begin{equation}
    \phi^{k+1}=\argmin\limits_{\phi}E_1(\phi)=E(\phi,\vec{w}^k),
\end{equation}
\begin{equation}
    \begin{cases}
        \vec{w}^{k+1}=\argmin\limits_{\vec{w}}E_2(\vec{w})=E(\phi^{k+1},\vec{w})\\
        s.t. \text{ }|\vec{w}|=1
    \end{cases}.
\end{equation}

To solve the sub-problem (24), we can derive the following evolution equation of $\phi$ via standard variational methods \cite{47},

\begin{equation}\label{del_phi}
    \frac{\partial\phi(x,t)}{\partial t}=
    \left. \begin{cases}-\gamma g(x)|\vec{w}^k(x)| \delta_\varepsilon'(\phi(x,t)) +\alpha g(x)\delta_\varepsilon(\phi(x,t))+\mu\Delta\phi(x,t) \\
    +2\beta h_\varepsilon'(\phi(x,t))\vec{w}^k(x)\cdot\int_\Omega e^{-\frac{|x-y|^2}{d^2}}\vec{w}^k(y)h_\varepsilon(\phi(y,t))dy\\
    +\mu( \nabla\cdot\vec{b}^k(x)-\nabla\cdot\vec{w}^k(x))\\
    \end{cases}\right \},
\end{equation}

The initial condition and boundary condition are as below,

\begin{equation}
    \begin{cases}
        \phi^{k+1}(x)=\phi^{k}(x) & x\in\Omega\cup\partial\Omega\\
        \nabla\phi(x,t)\cdot\vec{n}=(\vec{w}^k(x)-\vec{b}^k(x))\cdot\vec{n} & x\in\partial\Omega,t\in[0,T]
    \end{cases},
\end{equation}

\noindent where,

\begin{equation}
    h^{\prime}_\varepsilon(\phi(x))=\delta_\varepsilon(\phi(x)+l)(1-H_\varepsilon(\phi(x)-l))-H_\varepsilon(\phi(x)+l)\delta_\varepsilon(\phi(x)-l).
\end{equation}

\begin{equation}
\delta_{\varepsilon}^{\prime}(\phi)=\left\{\begin{array}{cc}{-\frac{\pi}{2 \varepsilon^{2}} \sin \left(\frac{\pi \phi}{\varepsilon}\right)} & {|\phi| \leq \varepsilon} \\ {0} & {|\phi|>\varepsilon}\end{array},\right.
\end{equation}

With the Heaviside function originally adopted in \cite{16}, the newly introduced component $ \delta'_{\varepsilon}(\phi)$ in the Split Bregman algorithm may be excessively smoothed. Furthermore, as the SR is an edge-based model and the repelling force is local, smoothing $H(\phi)$ over the entire image causes the repelling force to propagate across the image, resulting in unnecessary instability. With the new choice of Heaviside function, the smoothing effect is restricted only to a narrow band of width $2\varepsilon$ surrounding the contour which in practice stabilizes contour evolution.

Next, we approximate the time derivative of $\phi(x,t)$ as $\frac{\partial\phi(x,t)}{\partial t}=\frac{\phi^{k+1}(x)-\phi^k(x)}{\tau}$ where $\tau$ is the time step. Rearranging (\ref{del_phi}), we get the following equation,

\begin{equation}\label{new_del_phi}
    (1-\tau\mu\Delta)\phi^{k+1}(x) = \phi^k(x) + \tau \left. \begin{cases}-\gamma g(x)|\vec{w}^k(x)| \delta_\varepsilon'(\phi^k(x)) +\alpha g(x)\delta_\varepsilon(\phi^k(x))\\
    +2\beta h_\varepsilon'(\phi^k(x))\vec{w}^k(x)\cdot\int_\Omega e^{-\frac{|x-y|^2}{d^2}}\vec{w}^k(y)h_\varepsilon(\phi^k(y))dy\\
    +\mu \nabla\cdot(\vec{b}^k(x)-\vec{w}^k(x))\\
    \end{cases}\right\}.
\end{equation}

Using $F^k(x)$ to represent the r.h.s. of (\ref{new_del_phi}), we can derive the following by introducing FFT,

\begin{equation}\label{FFT}
    \mathbb{F}(1-\tau\mu\Delta)\mathbb{F}(\phi^{k+1}(x)) = \mathbb{F}(F^k),
\end{equation}

\noindent where,

\begin{equation}\label{FFT2}
    \mathbb{F}(1-\tau\mu\Delta) = 1-\tau\mu(2\cos{z^1_{l_1}}+2\cos{z^2_{l_2}-4}),
\end{equation}

\noindent for $z^1_{l_1}=\frac{2(l_1-1)\pi}{M}$, $z^2_{l_2}=\frac{2(l_2-1)\pi}{N}$, $l_1=1,2,...,M$, $l_2=1,2,...,N$, $M$ and $N$ are the row and column numbers of the image. The iterative formula of $\phi(x)$ can thus be derived as follows,

\begin{equation}\label{FFT_phi}
    \phi^{k+1}(x)=\mathbb{R}\left(\mathbb{F}^{-1}\left(\frac{\mathbb{F}(F^k(x))}{\mathbb{F}(1-\tau\mu\Delta)}\right)\right).
\end{equation}

For the sub-problem (25), if $|\vec{w}(x)|\neq 0$, we can obtain the corresponding Euler-Lagrange equation of $\vec{w}(x)$ as,

\begin{equation}\label{w_subprob}
    \begin{cases}
        \gamma g(x)\delta_\varepsilon(\phi^{k+1}(x))\frac{\vec{w}(x)}{|\vec{w}(x)|}-2\beta h_\varepsilon(\phi^{k+1}(x))\int_\Omega e^{-\frac{|x-y|^2}{d^2}}\vec{w}(y)h_\varepsilon(\phi^{k+1}(y,t))dy\\
    +\mu(\vec{w}(x)-\nabla\phi^{k+1}(x)-\vec{b}^k(x))=0\\
    s.t. \text{ }|\vec{w}(x)|=1
    \end{cases}.
\end{equation}

However, since the second term in (\ref{w_subprob}) contains the integral of $\vec{w}(y)$, it is not straightforward to construct the iterative scheme for $\vec{w}^k$. An approximation formula with projection is designed in the next section to address this issue.

\section{Discretization and Iterative Scheme}

For the next step in solving (\ref{w_subprob}), we devise the discretization of the continuous derivatives. Let the spatial step be 1 and time step be $\tau$, and the discrete coordinates for the pixel $(i,j)$ be $x_{i,j}=(x_{1i},x_{2j})$ where $i=0,1,2,...,M+1$, $j=0,1,2,...,N+1$ , we get $\phi_{i,j}=\phi(x_{1i},x_{2j})$. Let the other variables take similar forms. With the first order finite difference approximation, we can obtain the discrete gradient, Laplacian, and divergences respectively as,

\begin{equation}
\begin{array}{l}
    \nabla\phi_{i,j}=
    \begin{bmatrix}
        \phi_{i+1,j}-\phi_{i,j}\\
        \phi_{i,j+1}-\phi_{i,j}
    \end{bmatrix},\\
    \Delta\phi_{i,j}=\phi_{i-1,j}+\phi_{i,j-1}+\phi_{i+1,j}+\phi_{i,j+1}-4\phi_{i,j}.
\end{array}
\end{equation}

\begin{equation}
\begin{array}{l}
    \nabla\cdot\vec{w}_{i,j}=(\vec{w}_{1i,j}-\vec{w}_{1i-1,j})+(\vec{w}_{2i,j}-\vec{w}_{2i,j-1}),\\ \nabla\cdot\vec{b}_{i,j}=(\vec{b}_{1i,j}-\vec{b}_{1i-1,j})+(\vec{b}_{2i,j}-\vec{b}_{2i,j-1}),
\end{array}
\end{equation}

The first order time derivative of $\phi_{i,j}$ can be approximated as $\frac{\partial\phi_{i,j}}{\partial t}=\frac{\phi_{i,j}^{k+1}-\phi_{i,j}^{k}}{\tau}$. Therefore, from (\ref{FFT_phi}), a semi-implicit iterative scheme can be designed for $\phi_{i,j}^{k+1,s+1}$ where $s=0,1,2,...,S$, such that,

\begin{equation}\label{phi_numeric}
\begin{array}{l}
\phi^{k+1,0}_{i,j}=\phi^{k}_{i,j},\\

\phi^{k+1,s+1}_{i,j}(x)=\mathbb{R}\left(\mathbb{F}^{-1}\left(\frac{\mathbb{F}(F^{k,s}(x))}{\mathbb{F}(1-\tau\mu\Delta)}\right)\right)_{i,j},\\

F^{k,s}_{i,j}(x)= \phi^{k,s}_{i,j}(x) + \tau \left. \begin{cases}-\gamma g_{i,j}(x)|\vec{w}^k_{i,j}(x)| \delta_\varepsilon'(\phi^{k,s}_{i,j}(x)) +\alpha g_{i,j}(x)\delta_\varepsilon(\phi^{k,s}_{i,j}(x))\\
    +2\beta h_\varepsilon'(\phi^{k,s}_{i,j}(x))\vec{w}^k_{i,j}(x)\cdot\vec{v}^{k,s}_{i,j}\\
    +\mu \nabla\cdot(\vec{b}^k_{i,j}(x)-\vec{w}^k_{i,j}(x))\\
    \end{cases}\right\}.

%\frac{\phi_{i,j}^{k+1,s+1}-\phi_{i,j}^{k+1,s}}{\tau}=
%\left.\begin{cases}
%    -2\gamma g_{i,j}|\vec{w}^k_{i,j}|\delta_{\varepsilon}^{\prime}(\phi_{i, j}^{k+1, s})+\alpha g_{i,j}\delta_\varepsilon(\phi_{i,j}^{k+1,s})\\
%    +\mu(\phi_{i-1,j}^{k+1,s+1}+\phi_{i,j-1}^{k+1,s+1}+\phi_{i+1,j}^{k+1,s}+\phi_{i,j+1}^{k+1,s}-4\phi_{i,j}^{k+1,s+1})\\
%    +2\beta h^{\prime}_\varepsilon(\phi_{i,j}^{k+1,s})\vec{w}_{i,j}^k\cdot\vec{v}_{i,j}^k
%    +\mu(\nabla\cdot\vec{b}_{i,j}^k-\nabla\cdot\vec{w}_{i,j}^k) 
%\end{cases}\right\},
\end{array}
\end{equation}

\noindent until $\frac{\left\|\phi^{k+1, s+1}-\phi^{k+1, s}\right\|}{\left\|\phi^{k+1, s}\right\|+10^{-6}} \leq T o l$. 

$\vec{v}_{i, j}^{k, s}=\left(\sum\limits_{p=-d}^{d} \sum\limits_{q=-d}^{d} e^{-\frac{\left(p^{2}+q^{2}\right)}{d^{2}}} \vec{w}_{i+p, j+q}^{k} h_{\varepsilon}\left(\phi_{i+p, j+q}^{k+1, s}\right)\right)$
\linebreak which is the discrete approximation of $\vec{v}^{k}(x)=\int_{\Omega} e^{-\frac{|x-y|^{2}}{d^{2}}} \vec{w}^{k}(y) h_{\varepsilon}(\phi(y, t)) d y$. $y$ denotes a point taken from a small window of size $2d \times 2d$ around point $x$. The repulsion from points further away is negligible, therefore we only check within a small window. Note that the initial and boundary conditions in (27) still hold.

Next, we will solve (\ref{w_subprob}) iteratively. By temporarily fixing $\vec{w}^{k+1, r}(y)$, we can design a concise approximate generalized soft thresholding formula. For abbreviation, let

\begin{equation}
\vec{v}^{k+1, r}(x)=\int_{\Omega} e^{-\frac{ | x-y ]^{2}}{d^{2}}} \vec{w}^{k+1, r}(y) h_{\varepsilon}\left(\phi^{k+1}(y)\right) d y,
\end{equation}

\noindent and $\vec{w}^{k+1,0}(y)=\vec{w}^{k}(y)$. For $r=0,1,2, \ldots, R$, since $|\vec{w}_{i,j}^{k+1,r}|=1$, the iterative formula for $\vec{w}^{k+1}$ from (25) can be written as,

\begin{equation}\label{w_iter}
\vec{\tilde{w}}_{i, j}^{k+1, r+1} \approx \frac{\mu \nabla\phi^{k+1}_{i,j} + \mu\vec{b}^{k}_{i,j} + 2\beta h_{\varepsilon}\left(\phi^{k+1}_{i,j}\right) \vec{v}^{k+1, r}_{i,j}}{\gamma g_{i,j} \delta_\varepsilon(\phi^{k+1}_{i,j}) + \mu},
\end{equation}

\begin{equation}\label{w_projection}
\vec{w}_{i, j}^{k+1, r+1}=\frac{\vec{\tilde{w}}_{i, j}^{k+1, r+1}}{\left|\vec{\tilde{w}}_{i, j}^{k+1, r+1}\right|}.
\end{equation}

In practice, a single iteration is often enough for computing (\ref{w_iter}).
Alternatively, we can directly use the soft thresholding formula to derive $\vec{w}^{k+1}$. For abbreviation, let

\begin{equation}
\vec{B}^{k+1}=\nabla\phi^{k+1}(x) + \vec{b}^{k} + \frac{2\beta}{\mu} h_{\varepsilon}\left(\phi^{k+1}(x)\right)\int_{\Omega} e^{-\frac{ | x-y ]^{2}}{d^{2}}} \vec{w}^{k}(y) h_{\varepsilon}\left(\phi^{k+1}(y)\right) d y.
\end{equation}

The formula for $\vec{w}_{i,j}^{k+1}$ is

\begin{equation}\label{w_numeric}
\vec{w}_{i,j}^{k+1}\approx \max (|\vec{B}^{k+1}_{i,j}|-\frac{\gamma}{\mu}g_{i,j}\delta_\varepsilon(\phi^{k+1}_{i,j}),0)\frac{\vec{B}^{k+1}_{i,j}}{|\vec{B}^{k+1}_{i,j}|}, 0\frac{\vec{0}}{|\vec{0}|} = \vec{0}.
\end{equation}

The same projection scheme as (\ref{w_projection}) is used afterwards. After $\phi_{i, j}^{k+1}, \vec{w}_{i, j}^{k+1}$ have been obtained, we can derive $\vec{b}_{i, j}^{k+1}$ directly from (23).

In summary, the Split-Bregman algorithm proposed in this section has four advantages. First, the simplified algorithm and the use of a projection scheme in place of the initialization step improves efficiency. Both the per-iteration time and the convergence time have been reduced as shown in the experiments section. Second, the memory requirement is reduced. For an image of size $M\times N$, the parameter $A$ in the AOS solution is size $2\times(M\times N)\times(M\times N)$. In the Split Bregman algorithm, the sizes of both $\vec{w}$ and $\vec{b}$ are $2\times(M\times N)$ only. Third, the evolution of the contour is stabilized by a semi-implicit FFT scheme and smoothing the Heaviside function only within the narrow-bands around the contours. These changes allow for bigger time steps and more lenient parameter tuning. Finally, the numerical solution is simplified. In (17), the convolution term containing $\nabla\phi$ is hyperbolic, which requires the upwind difference scheme. By substituting $\nabla\phi$ with the auxiliary variable $\vec{w}$ we can remove the need for complex discretization schemes. 

The proposed algorithm is summarized in Algorithm 1.

\begin{algorithm}
\begin{enumerate}[(1)]
\item Initialize\\
Calculate $g(x)$ using (1)\\
Initialize $\phi^{0}(x)$ as a signed distance function and set $\vec{w}^{0}=\nabla\phi^0, \vec{b}^{0}=\overrightarrow{0}$\\
Set penalty parameters\\
Set tolerance errors, time step and iterative steps\\
\item Iterations\\
 For $k$=0,1,2,...,K\\
\qquad For $s$=0,1,2,...,S\\
\qquad\qquad Calculate $\phi^{k+1,s+1}$ from (\ref{phi_numeric})\\
\qquad End for $s$ when (24) converges\\
\qquad Calculate $\vec{w}^{k+1}$ from (\ref{w_numeric})\\
%%\qquad For $l$=0,1,2,...,L\\
%%\qquad\qquad Calculate $\vec{w}^{k+1, l+1}$ from (32), (33)\\
%%\qquad End for $l$ when (21) converges\\
\qquad Calculate $\vec{b}^{k+1}$ from (23)\\
End for $k$ when (13) converges
\end{enumerate}
\caption{The Split Bregman algorithm for the Self-repelling Snake Model}
\end{algorithm}

With regards to parallelization, we can consider the $\phi$ and $w$ sub-problems separately. $w$ can be solved directly with an approximate soft thresholding formula and requires no iterations. $\phi$, on the other hand, is now solved with discrete FFT which has an abundance of pre-existing fast parallel implementations. 

Finally, it is worth mentioning that the $\phi$ sub-problem can be solved with an AOS scheme as well. In this case, the coefficient $A_l$ will be constant and LR decomposition will only need to happen once compared to once every iteration in the original AOS solution. Nonetheless, both the FFT and AOS schemes are strongly semi-implicit compared to Gauss-Seidel iterations, leading to the stability of the algorithm.

\section{Numerical Experiments}
\subsection{Experimental Results}
The experiments below demonstrate that the Split Bregman solution of the SR model can successfully prevent contour splitting and merging. The qualitative performance is comparable to the original algorithm while the time to reach convergence is shortened and the memory usage is reduced. Two practical applications are showcased as well as the adaptation to 3D. All experiments are performed on the PC (Intel(R)  Core  (TM)  i7-7700  CPU  @  3.60GHz  3.60  GHz; 16.0  GB memory). The segmentation program is written in Matlab and runs in Matlab environment R2021a.

\begin{figure}[H]\centering
\begin{subfigure}{.32\textwidth}
\centering
\includegraphics[width=\linewidth]{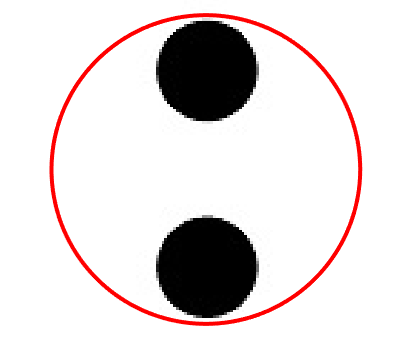}
\caption{}
\end{subfigure}
\begin{subfigure}{.32\textwidth}
\centering
\includegraphics[width=\linewidth]{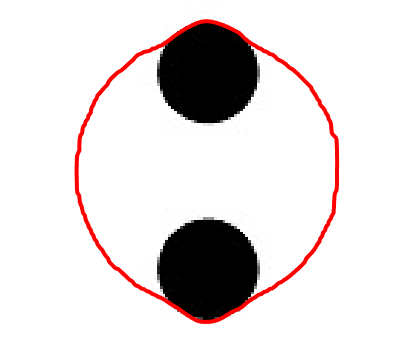}
\caption{}
\end{subfigure}
\begin{subfigure}{.32\textwidth}
\centering
\includegraphics[width=\linewidth]{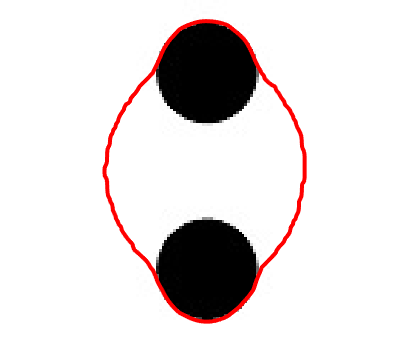}
\caption{}
\end{subfigure}\\
\begin{subfigure}{.32\textwidth}
\centering
\includegraphics[width=\linewidth]{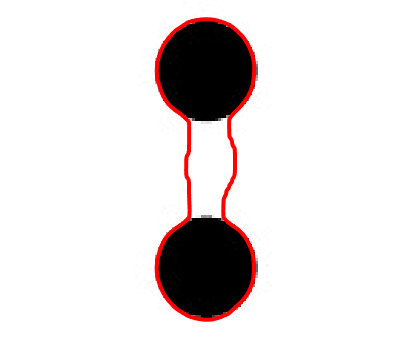}
\caption{}
\end{subfigure}
\begin{subfigure}{.32\textwidth}
\centering
\includegraphics[width=\linewidth]{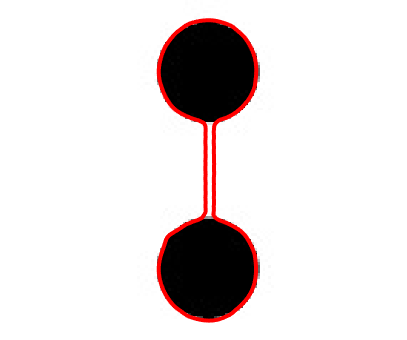}
\caption{}
\end{subfigure}
\begin{subfigure}{.32\textwidth}
\centering
\includegraphics[width=\linewidth]{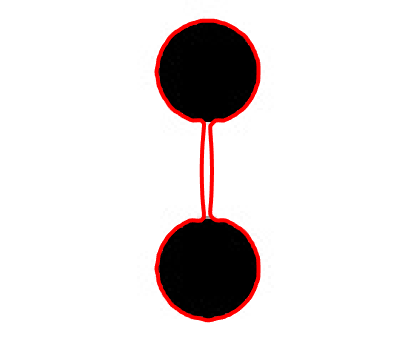}
\caption{}
\end{subfigure}
\caption{Segmentation of two circles with the Split Bregman algorithm, with image taken from \cite{16}. (a)-(f) are the segmentation steps via the Split Bregman algorithm, (f) is the result of the AOS algorithm for comparison.  $\alpha=4$, $\gamma=4$, $\beta=0.2$, $\mu=8$, $l=1$, $d=5$, $window=5\times 5$, $S=3$, $\varepsilon=1$, $\tau=.1$, $Tol=10^{-5}$.}
\end{figure}

\begin{figure}[H]\centering
\centering
\includegraphics[width=0.5\linewidth]{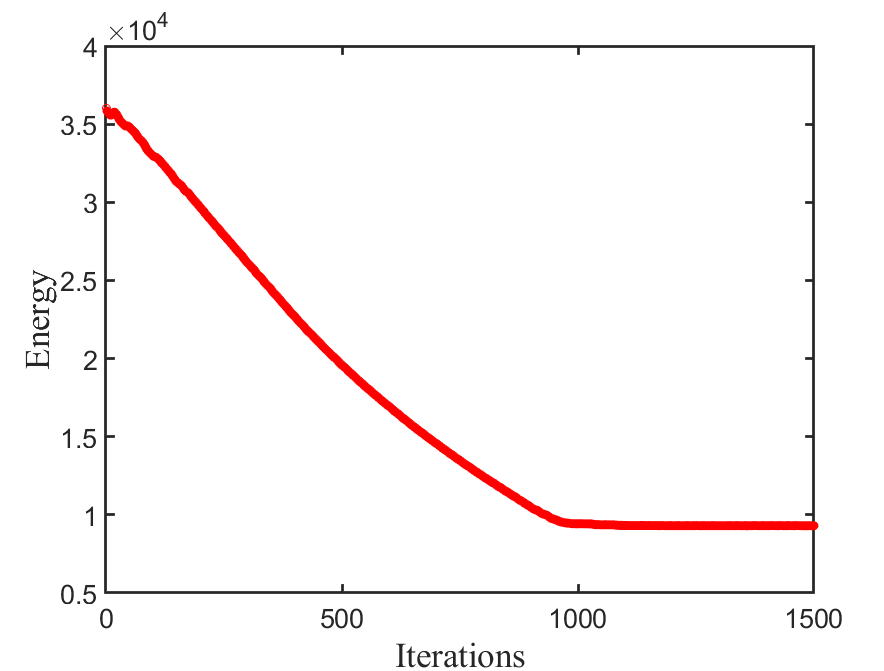}
\caption{Convergence graph for the Split Bregman algorithm in the segmentation of two circles experiment. Convergence was reached at step 1189.}
\end{figure}

In Figure 1, we can see that contour splitting is prevented and the topology is successfully preserved. Figure 1 (e) and (f) show that comparable results were obtained from the Split-Bregman algorithm and the AOS algorithm. Convergence was reached by step 1189 in the Split Bregman case and by approximately 500 in the AOS case. However, the Split Bregman algorithm (written in Matlab) took 6.10s while the original AOS algorithm (written in C) took 36.18s. This shows that the per-iteration time has been significantly reduced for the Split Bregman algorithm, resulting in shorter convergence times.

Adjustments can be made on the various parameters to improve segmentation quality. Parameter $\alpha$ controls the outwards or inwards driving force, $\gamma$ dictates the geodesic length, $\beta$ weights the repelling force, and $\mu$ weights the constraint. An excessively large $\beta$ causes the contour to become unstable, as the repulsive force is a nonlocal term. However, increasing $\beta$ and decreasing the window size narrows the gap between the contours. Typically, the window size is $5\times 5$ or $7\times 7$ as according to \cite{16}. A smaller time step $\tau$ increases stability. Increasing $\varepsilon$ improves the smoothness of the contour but lowers the effectiveness of topology preservation, as it smooths out the repulsive force. In practice, we can start from the same set of parameters and only make minor changes as appropriate.

\begin{figure}[H]\centering
\begin{subfigure}{.24\textwidth}
\centering
\includegraphics[width=\linewidth]{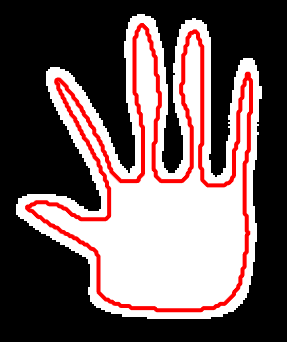}
\caption{}
\end{subfigure}
\begin{subfigure}{.24\textwidth}
\centering
\includegraphics[width=\linewidth]{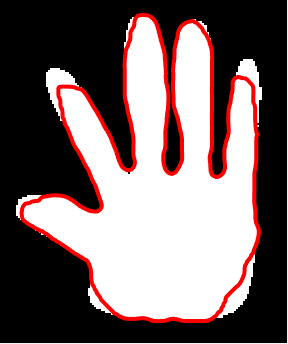}
\caption{}
\end{subfigure}
\begin{subfigure}{.24\textwidth}
\centering
\includegraphics[width=\linewidth]{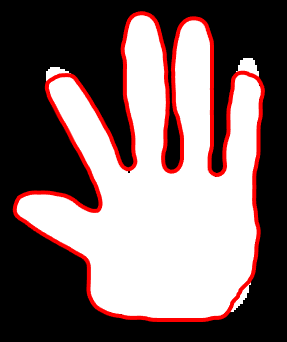}
\caption{}
\end{subfigure}
\begin{subfigure}{.24\textwidth}
\centering
\includegraphics[width=\linewidth]{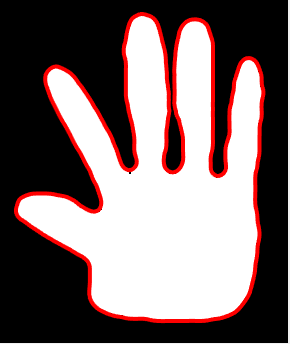}
\caption{}
\end{subfigure}

\caption{Segmentation of synthetic hand with the Split Bregman algorithm, image taken from \cite{16}. $\alpha=4.5$, $\gamma=5$, $\beta=0.3$, $\mu=8$, $l=1$, $d=4$, $window=5\times 5$, $S=3$, $\varepsilon=1$, $\tau=.05$, $Tol=10^{-6}$.}
\end{figure}

In Figure 3, contour merging is prevented as the fingers of the hand remain separate. In the basic GAC model, the proximity of the contours would cause them to merge despite there being a detected edge, because it reduces the total geodesic length. Note that contours was initialized with basic binary thresholding to increase efficiency.

% \begin{figure}[H]
% \begin{center}

% \begin{subfigure}{.33\textwidth}
% \centering
% \includegraphics[width=\linewidth]{brain1.png}
% \caption{}
% \end{subfigure}
% \begin{subfigure}{.33\textwidth}
% \centering
% \includegraphics[width=\linewidth]{brain2.png}
% \caption{}
% \end{subfigure}
% \begin{subfigure}{.33\textwidth}
% \centering
% \includegraphics[width=\linewidth]{brain3.png}
% \caption{}
% \end{subfigure}
% \begin{subfigure}{.33\textwidth}
% \centering
% \includegraphics[width=\linewidth]{brain5.png}
% \caption{}
% \end{subfigure}
% \end{center}
% \caption{Segmentation of a brain, image taken from \cite{16}. (a) is the original image, (b-c) uses the new Heaviside function with $\alpha=10$, $\gamma=15$, $\beta=2$, $\mu=5$, $l=1$, $d=4$, $window=7\times 7$, $S=5$, $\varepsilon=1$, $\tau=.1$, $Tol=10^{-6}$. (d) uses the original Heaviside function with the same parameters.}
% \end{figure}

% Figure 4 compares the effect of two different Heaviside functions. The advantage of the new Heaviside formulation lies in the stabilization of the repelling term, which makes the algorithm more robust. In the experiment in Figure 3, the amount of repulsion was set to very high through $\beta$. Yet, it still did not disturb the smoothness of the contour nor cause the loss of necessary details. With the same set of parameters and the original Heaviside function, the repulsive force was dissipated and stability could not be maintained.

\begin{figure}[H]
\begin{subfigure}{.32\textwidth}
\centering
\includegraphics[width=\linewidth]{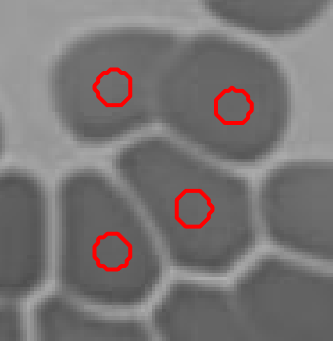}
\caption{}
\end{subfigure}
\begin{subfigure}{.32\textwidth}
\centering
\includegraphics[width=\linewidth]{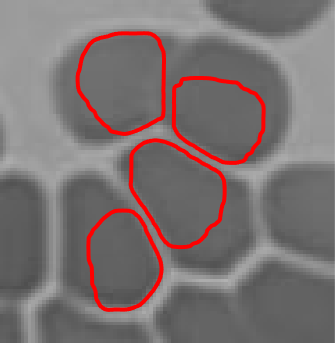}
\caption{}
\end{subfigure}
\begin{subfigure}{.32\textwidth}
\centering
\includegraphics[width=\linewidth]{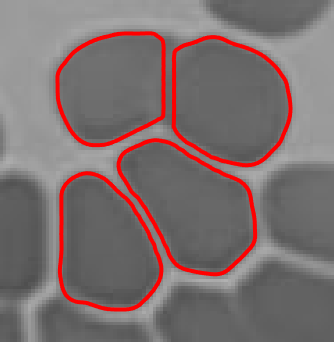}
\caption{}
\end{subfigure}
\caption{Segmentation of cells (image taken from \cite{46}). $\alpha=4.5$, $\gamma=5$, $\beta=0.25$, $\mu=9$, $l=1$, $d=4$, $window=5\times 5$, $S=3$ $\varepsilon=1$, $\tau=.02$, $Tol=10^{-6}$.}
\end{figure}

\begin{figure}[H]
\begin{subfigure}{.32\textwidth}
\centering
\includegraphics[width=\linewidth]{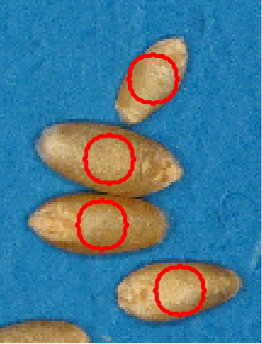}
\caption{}
\end{subfigure}
\begin{subfigure}{.32\textwidth}
\centering
\includegraphics[width=\linewidth]{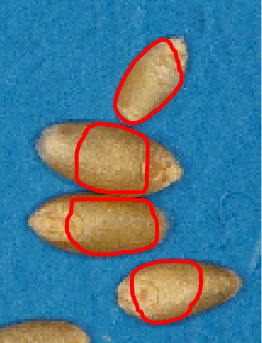}
\caption{}
\end{subfigure}
\begin{subfigure}{.32\textwidth}
\centering
\includegraphics[width=\linewidth]{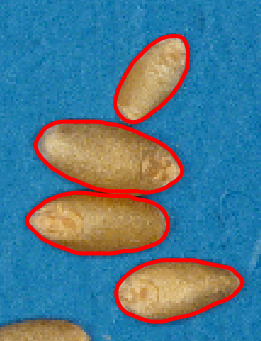}
\caption{}
\end{subfigure}
\caption{Segmentation of wheat grains. $\alpha=5$, $\gamma=5$, $\beta=0.25$, $\mu=8$, $l=1$, $d=4$, $window=5\times 5$, $S=3$ $\varepsilon=1$, $\tau=.02$, $Tol=10^{-6}$.}
\end{figure}

Two notable examples of practical applications of the algorithm are adhesive cell segmentation and grain segmentation. As seen in Figure 4 and Figure 5, the repulsive term prevents the contours of cells and grains from merging. The centers of the cells and grains can be detected via k-means clustering or detector filters such as the circle Hough Transform or the Laplacian of Gaussian \cite{45}. Since the topology is maintained, the number of entities will remain the same.

\begin{figure}[H]
\centering
\begin{subfigure}{.32\textwidth}
\centering
\includegraphics[width=\linewidth]{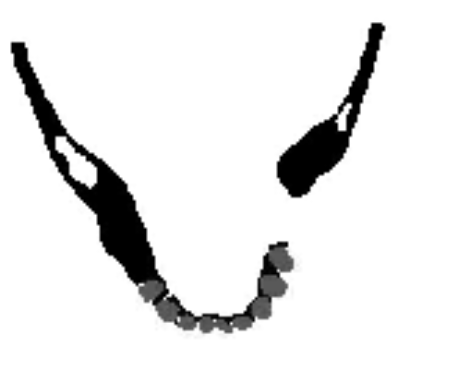}
\caption{}
\end{subfigure}
\begin{subfigure}{.32\textwidth}
\centering
\includegraphics[width=\linewidth]{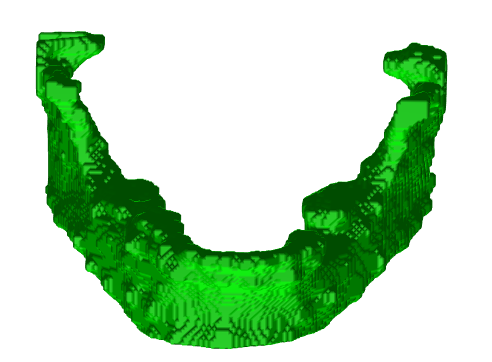}
\caption{}
\end{subfigure}
\begin{subfigure}{.32\textwidth}
\centering
\includegraphics[width=\linewidth]{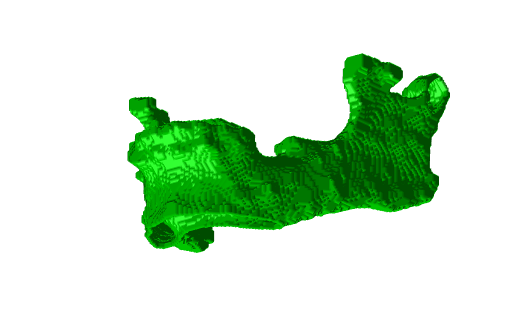}
\caption{}
\end{subfigure}
\caption{Segmentation of a human mandible from 105 CT scan images.(a) is one of the CT images, (b)(c) are different views of the segmentation contour. $\alpha=4$, $\gamma=5$, $\beta=0.2$, $\mu=3$, $l=1$, $d=4$, $window=5\times 5$, $S=5$ $\varepsilon=1$, $\tau=.05$, $Tol=10^{-6}$.}
\end{figure}

The algorithm can also be extended to 3D, as seen in Figure 6. The segmentation contour is generated from 105 CT scan images. The property of topology preservation prevents the splitting and merging of 3D components.

\section{Conclusions}
By introducing an intermediate variable and the Bregman iterative parameter, the Self-repelling Snake model can be solved through the Split-Bregman method. The problem is divided into two sub-problems that are solved with FFT and an approximate soft thresholding formula. A projection scheme is implemented instead of resorting to frequent re-initialization of the signed distance function. As a result, the new algorithm is able to maintain stability while simplifying computations, leading to shorter convergence time and reduced memory requirement. The algorithm is applicable to image segmentation problems where topology is a prior, e.g. adhesive cell segmentation, grain segmentation, 3D segmentation of medical imagery, etc. In future works, we will explore more 3D applications of the algorithm.

\section*{Acknowledgements}
Special thanks to Dr. C. Le Guyader (Universit\'e
de Rouen, France) for sharing her source code for the AOS solution of SR and Phobos Consulting for the grain imagery. Many thanks to the reviewers for their valuable comments and suggestions.

%%\section*{References}
\bibliography{article}
\end{document}